\documentclass[letterpaper, 10 pt, journal, twoside]{IEEEtran}

\usepackage{hyperref}

\hypersetup{
    colorlinks=true,
    linkcolor=blue,
    filecolor=magenta,      
    urlcolor=blue,
}

\newcommand*{\change}{\textcolor{black}}

\usepackage{graphics}           
\usepackage{times}              
\usepackage{amsmath}            
\usepackage{amssymb}            
\usepackage{graphicx}
\usepackage{algorithm}
\usepackage[noend]{algpseudocode}
\usepackage{booktabs}
\usepackage{color}
\definecolor{instructioncolor}{rgb}{.5,.5,.5}

\usepackage[font=small]{caption}

\def\secref#1{Sec.~\ref{#1}}
\def\figref#1{Fig.~\ref{#1}}
\def\tabref#1{Tab.~\ref{#1}}
\def\eqref#1{Eq.~(\ref{#1})}

\makeatletter
\usepackage{xspace}
\DeclareRobustCommand\onedot{\futurelet\@let@token\@onedot}
\def\@onedot{\ifx\@let@token.\else.\null\fi\xspace}
 
\def\ie{i.e\onedot}

\def\etal{{et al}\onedot}
\makeatother

\def\etalcite#1{\etal~\cite{#1}}

\usepackage{array}
\newcolumntype{L}[1]{>{\raggedright\let\newline\\\arraybackslash\hspace{0pt}}m{#1}}
\newcolumntype{C}[1]{>{\centering\let\newline\\\arraybackslash\hspace{0pt}}m{#1}}
\newcolumntype{R}[1]{>{\raggedleft\let\newline\\\arraybackslash\hspace{0pt}}m{#1}}

\newcommand{\RR}{\mathbb{R}}

\renewcommand{\b}[1]{\mbox{\boldmath$#1$}}

\newcommand{\set}[1]{\mathcal{#1}} 	

\renewcommand{\vec}[1]{{\b #1}} 
\renewcommand{\v}[1]{{\b #1}} 

\usepackage{subfigure}
\usepackage{multirow}
\usepackage{rotating}  

\title{\LARGE \bf OverlapTransformer: An Efficient and \change{Yaw-Angle-Invariant} \\ Transformer Network for LiDAR-Based Place Recognition}

\author{Junyi Ma, Jun Zhang, Jintao Xu, Rui Ai, Weihao Gu, Xieyuanli Chen$^*$ 
\thanks{Manuscript received: February 21, 2022; Revised: April 26, 2022; Accepted: May 19, 2022. 
  This paper was recommended for publication by Editor Javier Civera upon evaluation of the Associate Editor and Reviewers' comments.}

  \thanks{
  J. Ma is with the Beijing Institute of Technology.
  J. Ma, J. Zhang, J. Xu, R. Ai and W. Gu are with HAOMO.AI Technology Co., Ltd.}
  \thanks{$^*$corresponding author email: chenxieyuanli@hotmail.com}
  \thanks{This work has partially been funded by the HAOMO.AI Technology Co. Ltd., and by the Chinese Scholarship Committee.
  }%
  \thanks{Digital Object Identifier (DOI): see top of this page.}
}

\begin{document}
\maketitle

\IEEEpeerreviewmaketitle

\markboth{IEEE Robotics and Automation Letters. Preprint Version. Accepted May, 2022}
{Ma \MakeLowercase{\textit{et al.}}: OverlapTransformer: An Efficient and Yaw-Angle-Invariant Transformer Network for LiDAR-Based Place Recognition}

\begin{abstract}

Place recognition is an important capability for autonomously navigating vehicles operating in complex environments and under changing conditions. It is a key component for tasks such as loop closing in SLAM or global localization. In this paper, we address the problem of place recognition based on 3D LiDAR scans recorded by an autonomous vehicle. \change{We propose a novel lightweight neural network exploiting the range image representation of LiDAR sensors to achieve fast execution with less than 2\,ms per frame. We design a yaw-angle-invariant architecture exploiting a transformer network, which boosts the place recognition performance of our method.} We evaluate our approach on the KITTI and Ford Campus datasets. The experimental results show that our method can effectively detect loop closures compared to the state-of-the-art methods and generalizes well across different environments. \change{To evaluate long-term place recognition performance, we provide a novel dataset containing LiDAR sequences recorded by a mobile robot in repetitive places at different times.
The implementation of our method and dataset are released here: \url{https://github.com/haomo-ai/OverlapTransformer}}

\end{abstract}


\section{Introduction}
\label{sec:intro}

Place recognition plays an essential role for autonomously navigating systems. In contrast to camera-based place recognition~\cite{arandjelovic2016netvlad, hausler2021patch, lowry2016tro, vysotska2016ral, vysotska2019ral}, LiDAR-based place recognition~\cite{chen2020rss, kim2018scan, uy2018pointnetvlad, wang2020lidar} is comparably robust to day-and-night light changes and different weather conditions. Therefore, LiDARs are an attractive sensing modality that can be used for autonomous driving in outdoor large-scale environments. 
LiDAR-based place recognition is the task to determine if the LiDAR sensor is currently in a place that has been visited before by comparing the current LiDAR observations with the map or a database of formerly taken observations.
It supports tasks such as simultaneous localization and mapping (SLAM) to find loop closure candidates and global localization to obtain an initial guess of the robot's position.

\begin{figure}
  \centering
  \includegraphics[width=1\linewidth]{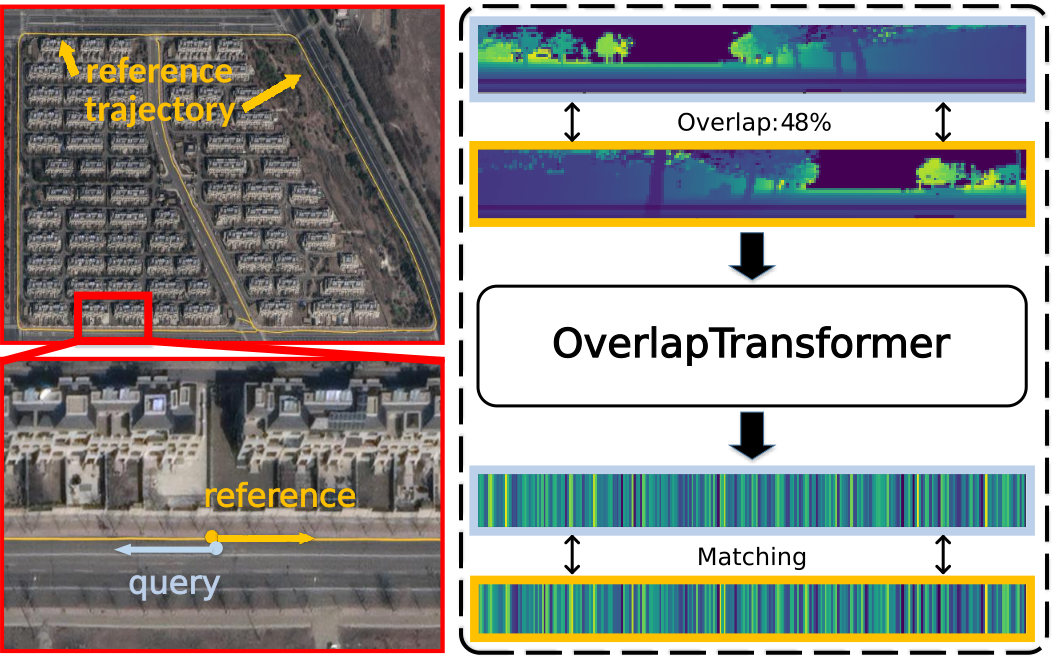}
  \caption{Query scan (blue) and reference scan (orange) with adjacent locations but opposite viewpoints in our novel Haomo dataset.
    Our OverlapTransformer is able to generate \change{yaw-angle-invariant} global descriptors with only range images, which is robust to viewpoint changing for place recognition.}
  \label{fig:motivation}
  \vspace{-0.5cm}
\end{figure}

In this paper, we propose a novel place recognition method utilizing range images produced by 3D LiDARs installed on an autonomous vehicle.
The range image is a natural representation of a single 3D scan from a rotating LiDAR sensor such as Velodyne or Ouster sensors. It is a compact representation and is especially suitable for online tasks such as online SLAM~\cite{chen2019iros}, loop closing~\cite{chen2021auro,chen2020rss}, or localization~\cite{chen2020iros,chen2021icra} due to its image-like structure. 
Instead of using handcrafted descriptors~\cite{he2016iros,roehling2015iros,steder2010irosws},
we propose a new transformer neural network to extract \change{yaw-angle-invariant} global descriptors from LiDAR scans. 
We apply a similar overlap concept as described in OverlapNet~\cite{chen2021auro,chen2020rss} to supervise the network learning and to estimate the similarity between pairs of scans.
Different to OverlapNet, which exploits multiple cues such as normal, intensity, and semantic information as the network input, in this work we only use the depth information of the range image to achieve faster online performance and make the approach easier to generalize. 
\change{Due to the yaw-angle-invariant design of the network, our approach can recognize places even when the vehicle drives in different directions, such as the reverse loop in our new Haomo dataset illustrated in~\figref{fig:motivation}.}

The main contribution of this paper is a lightweight transformer neural network that exploits only depth information of range images to achieve place recognition. Our approach is very fast to execute and at the same time yields very good recognition results. \change{Based on the attention mechanism of the Transformer~\cite{vaswani2017nips} and the NetVLAD head~\cite{arandjelovic2016netvlad}, our proposed OverlapTransformer compresses LiDAR range images into global descriptors. We build the architecture of our OverlapTransformer to ensure that each descriptor is yaw-angle-invariant, which makes our method robust to viewpoint changes.} 
We train the proposed OverlapTransformer only on a part of the KITTI dataset and evaluate it on both, KITTI and Ford Campus datasets with the loop closure metric in line with OverlapNet~\cite{chen2021auro}.
Besides, we recorded and released a new dataset, which contains three different challenges including place recognition for long time spans, reverse driving, and different appearance scenes to evaluate different methods.

In sum, we make the claims that our approach is able to
(i)~detect loop closure candidates for SLAM using only LiDAR data without any other  information, and generalize well into the different environments without fine-tuning,
(ii)~achieve long-term place recognition on our Haomo dataset in outdoor large-scale environments with a different LiDAR sensor,
\change{
(iii)~recognize places with changing viewpoints exploiting the proposed yaw-angle-invariant descriptors up to potentially discretization errors,
(iv)~run faster than most state-of-the-art place recognition methods.}

\section{Related Work}
\label{sec:related}

Place recognition is a common topic in computer vision and robotics with a large number of scientific work proposed using RGB images~\cite{jegou2010cvpr,arandjelovic2016netvlad,hausler2021patch,vysotska2016ral,vysotska2019ral}. We refer more image-based methods to survey by Lowry~\cite{lowry2016tro} and focus our discussion here more on 3D LiDAR-based approaches.

Due to the high accuracy of the range information and illumination invariance, LiDAR-based place recognition has attracted attention in the field of autonomous vehicles.
For example, He \etal~\cite{he2016iros} propose M2DP, which projects point cloud to multiple planes and combines descriptors from different planes to generate the global signature.
R\"ohling \etal~\cite{roehling2015iros} count the height of point cloud to generate the histogram-based 1-D global descriptor for fast retrieval.
Scan Context (SC) proposed by Kim \etal~\cite{kim2018scan} encodes the maximum height of point cloud in different bins to generate the 2D global descriptor for high discrimination, but leads to an increased computational matching time.
In contrast to SC using only geometric information, Cop \etal~\cite{cop2018delight} propose Delight to encode intensity-reading of LiDAR into a group of histograms for comprehensive utilization of intensity and geometric information.
Inspired by Cop~\etal, Wang \etal~\cite{wang2020intensity} extend SC by also exploiting both geometry and intensity, which outperforms geometric-only descriptors with the same space division method as used in SC.
Recently, Wang \etal~\cite{wang2020lidar} propose LiDAR Iris exploiting the Fourier transform to generate a binary signature image. It firstly generates LiDAR iris images by expanding the bird-eye view of the LiDAR scan into an image strip. Then, it applies Fourier transform on LiDAR iris images and solves the spatial place recognition in the frequency domain.

With the development of deep learning, more learning-based approaches are utilized for place recognition divided into two groups, local feature-based methods, and global descriptor-based methods.
Local feature-based approaches often have two steps, first extracting local features from the LiDAR scans and then recognizing places based on the extracted local features.
For example, Dube \etal~\cite{dube2017segmatch} propose SegMatch to firstly segment the filtered point cloud into sets of point clusters and then use features encoded by a CNN on such clusters to find place matches.
Based on SegMatch, Vidanapathirana \etal~\cite{vidanapathirana2021locus} propose Locus, which uses higher-order pooling along with a non-linear transformation to aggregate multi-level features and generate a fixed-length global descriptor for place recognition. 
LPD-Net by Liu \etal~\cite{liu2019lpd} uses ten types of local features from raw point cloud clusters as input of a graph neural network to generate global descriptors.
There are other learning-based approaches directly generating global descriptors on LiDAR scans for place recognition. 
For example, PointNetVLAD by Uy \etal~\cite{uy2018pointnetvlad} firstly uses PointNet~\cite{qi2017pointnet} to map laser points into a higher dimension and then uses the NetVLAD architecture~\cite{arandjelovic2016netvlad} previously used for visual place recognition to generate global descriptors for LiDAR-based place recognition. 
\change{Komorowski \etal~\cite{Komorowski2021wacv} utilize sparse 3D convolutions based on MinkowskiEngine with pooling layers to generate descriptors.}
\change{In contrast, LoGG3D-Net by Vidanapathirana \etal~\cite{vid2022icra} uses a local consistency loss to improve the performance of the global descriptor.}
SOE-Net by Xia \etal~\cite{xia2021soe} combines the orientation-encoding module with PointNet to generate point-wise features, which are fed to a self-attention network to generate discriminative and compact global descriptors. 
\change{Different to SOE-Net, Zhou \etal~\cite{zhou2021ndt} propose NDT-Transformer, which transforms raw point cloud into NDT cells and uses the attention mechanism from Transformer~\cite{vaswani2017nips} to improve the representation ability. PPT-Net by Hui \etal~\cite{hui2021iccv} exploits the pyramid point transformer module to enhance the discrimination of local features and generates a global descriptor.}

\change{
Our method also directly generates global descriptors on LiDAR scans. Different from methods that use local point cloud maps~\cite{liu2019lpd,uy2018pointnetvlad,xia2021soe,zhou2021ndt,hui2021iccv}, our method only uses range images generated from single 3D LiDAR scans. This yields fast computations suitable for online operation and natural yaw-angle-invariance for better place recognition performance. Our method also uses the Transformer similar to the work by Zhou \etal~\cite{zhou2021ndt} to boost place recognition performance, but our method operates on range images instead of NDT cells.}

Most recently, there are also works exploiting semantic information for place recognition.
For example, Chen \etal~\cite{chen2020rss,chen2021auro} propose OverlapNet to exploit multiple cues generated from LiDAR scan, including depth, normal, intensity, and semantics for LiDAR-based loop closure detection and localization.
SGPR by Kong \etal~\cite{kong2020semantic} exploits the semantics and topological information of the raw point cloud and extracts the semantic graph representation with graph neural networks to find loop closures.
Li \etal~\cite{li2021ssc} use semantics to enhance SC and propose semantic scan context. 
Similarly, Cramariuc \etal~\cite{cramariuc2021semsegmap} utilize semantic information to improve the performance of SegMatch.
\change{In contrast to these semantic-enhanced methods, our approach only uses the raw depth information to achieve online performance, which enables our method easier to generalize to different environments and datasets collected by different LiDAR sensors.}

\begin{figure*}[!t]
  \centering
  \includegraphics[width=0.95\linewidth]{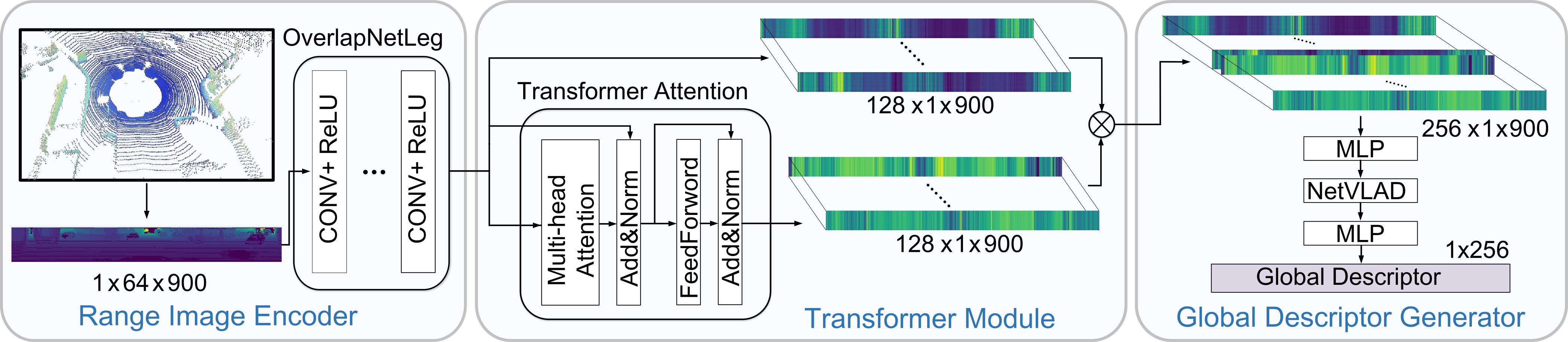}
  \caption{Pipeline overview of our proposed approach. 
    We take a 64-beam LiDAR as an example to introduce the dimensions of feature maps.
    The Range Image Encoder \change{(RIE)} compresses range images from the Lidar sensor to \change{yaw-angle-equivariant} feature maps.
    The Transformer Module \change{(TM)} concatenates feature maps from Range Image Encoder and discriminativeness-enhanced feature maps from Transformer Attention.
    \change{Global Descriptor Generator \change{(GDG)} generates yaw-angle-invariant descriptors exploiting the combination of MLP and NetVLAD.} 
    The generated 1-D global descriptors are used for fast retrieval of place recognition.}
  \label{fig:pipeline}
  \vspace{-0.2cm}
\end{figure*}

\section{Our Approach}
\label{sec:overlap_network}


The overview of our OverlapTransformer is depicted in~\figref{fig:pipeline}. The range image from raw point cloud is fed to an encoder, which is a modified OverlapNetLeg to extract features from range images (see~\secref{sec:range-image-encoder}). Then, the encoded feature volume is fed to the transformer module, where we utilize a transformer to embed the relative location of features and the global information across the whole range image (see~\secref{sec:AFT}). In the end, we use a \change{global descriptor generator with NetVLAD and multi-layer perceptrons (MLPs)} to compress the features and generate a \change{yaw-angle-invariant} 1-D global descriptor (see~\secref{sec:VLAD}). During training, we use the triplet loss with calculated overlap labels to better distinguish the positive and negative training examples (see~\secref{sec:loss}).

\subsection{Range Image Encoder}
\label{sec:range-image-encoder}

We use range images from LiDAR scans as input data. 
\change{The range image is an intermediate representation of LiDAR data obtained from a typical spinning mechanical LiDAR scanner with enough vertical resolution}. Given the LiDAR sensor parameters, there is a projection transformation between a point cloud and a range image.  
A point cloud~$\set{P}$ can be projected to a range image~$\set{R}, \Pi:~\RR^3 \mapsto \RR^2$, where each pixel contains one 3D point.
Each point~$\vec{p}_i = (x, y, z)$ is converted to image coordinates~$(u,v)$ by
\begin{align}
  \left( \begin{array}{c} u \vspace{0.0em}\\ v \end{array}\right) & = \left(\begin{array}{cc} \frac{1}{2}\left[1-\arctan(y, x) \pi^{-1}\right] w   \vspace{0.5em} \\
      \left[1 - \left(\arcsin(z r^{-1}) + \mathrm{f}_{\mathrm{up}}\right) \mathrm{f}^{-1}\right] h\end{array} \right), \label{eq:projection}
\end{align}
where~$r = ||\vec{p}||_2$ is the range,~$\mathrm{f} = \mathrm{f}_{\mathrm{up}} + \mathrm{f}_{\mathrm{down}}$ is the vertical field-of-view of the sensor, and~$w, h$ are the width and height of the resulting range image~$\set{R}$. 

The yaw rotation $\theta$ of a point cloud is \change{yaw-angle-equivariant} to the horizontal shift $s$ of the corresponding range image, and the following equations hold: 
\begin{align}
  \left( \begin{array}{c} u^{\prime} \vspace{0.0em}\\ v^{\prime} \end{array}\right) & = \left(\begin{array}{cc} u + s   \vspace{0.5em} \\
  v  \end{array} \right), \hspace{0.5em}
  s = \frac{1}{2}\left[1-\theta\pi^{-1}\right] w,
  \label{eq:projection2}
\end{align}
\begin{align}  
  \set{R} C_s = \Pi(R_\theta \set{P}).
\end{align}

The term $C_s$ represents the column shift of the range image $\set{R}$ by matrix right multiplication, and $R_\theta$ represents the yaw rotation matrix of the point cloud $\set{P}$.
\change{Note that, there might be discretization errors in pixel coordinates when generating the range images from point clouds.}

The range image with size of $h\times w\times 1$ is fed to the modified OverlapNetLeg, where $1$ refers to the one range channel, while $h, w$ are the height and width of the range image. Compared to OverlapNet~\cite{chen2021auro}, the convolution filters in our encoder only compress the range image in the vertical dimension but not the width dimension to avoid the discretization error to the \change{yaw equivariance}. Besides, there is no padding and dropout in our proposed architecture to keep the \change{yaw equivariance} for every intermediate feature generated by each network layer. We denote the output feature volume of the range image encoder as $\set{F} = \change{\text{RIE}}(\set{R})$, with the size of $1 \times w\times c $, and $C_s \set{F} = \change{\text{RIE}}(\Pi(R_\theta \set{P}))$ holds. The variable $c$ refers to the channel number of the encoded feature.
\vspace{-0.5em}
\subsection{Transformer Module}
\label{sec:AFT}

Inspired by NDT-transformer~\cite{zhou2021ndt}, we also exploit a transformer to extract more distinctive features for LiDAR place recognition. As shown in the middle of~\figref{fig:pipeline}, our attentional feature transformer is composed of three modules, multi-head self-attention (MHSA), feed-forward network (FFN), and layer normalization (LN). 
\change{Different from NDT-transformer, we use one transformer block in our transformer module to achieve high accuracy as well as efficiency.}

The MHSA can learn the relationship between features captured by a self-attention mechanism. We denote the feature volume extracted by the MHSA as $\set{A}$, and the self-attention mechanism of the transformer can then be fomulated as:
\change{
\begin{align}
  \set{A}&=\text{Attention}(Q, K, V) 
  =\text{softmax} \Big(\frac{Q {K}^{T}}{\sqrt{d_k}}\Big) V \vspace{0.5em} ,
 \label{eq:ScaledDotProductAttention}
\end{align}
where $\{Q, K, V\}$ are the query, key and value splits from the feature volumes generated by our range image encoder, and $d_k$ represents the dimension of splits.
}

\change{$\set{A}$ is then fed into the FFN and LN to generate the final attentional feature volume $\set{S}$, which can be calculated as:}
\begin{align}
  \set{S} &= \text{LN} ( \text{FFN}(\text{LN}(\text{Conc}(\set{F}, \set{A}))) + \text{LN}(\text{Conc}(\set{F}, \set{A}))),
 \label{eq:ScaledDotProductAttention}
\end{align}
where $\text{Conc}(\cdot)$ represents the concatenation operation across channels. In this way, a coarse feature~$\set{F}$ extracted by our range image encoder is upgraded to an attentional feature~$\set{S}$.

Here we further show that our transformer module is also \change{yaw-angle-equivariant}. 
It is obvious that concatenation across channels, linear transformations, ReLU function, and LN are \change{yaw-angle-equivariant}. Since the query, key, and value feature volumes in attention mechanism are split along the channel dimension, given a shifted feature generated by our range image encorder $C_s \set{F}$, we have:
\begin{align}
\text{Attention}&(C_s Q, C_s K, C_s V) \vspace{0.5em} \\
  &= \text{softmax}\Big(\frac{C_s Q (C_s K)^{T}}{\sqrt{d_k}}\Big)C_s V \vspace{0.5em} \\
  &= \text{softmax}\Big(\frac{C_s Q {K}^{T} {C_s}^{T}}{\sqrt{d_k}}\Big)C_s V \vspace{0.5em} \\
  &= C_s \text{softmax}\Big(\frac{Q {K}^{T} }{\sqrt{d_k}}\Big){C_s}^{T}C_s V \vspace{0.5em} \\
  &= C_s \set{A}.
 \label{eq:ScaledDotProductAttention}
\end{align}

Since FFN is applied to each element in the feature volume separately and identically~\cite{vaswani2017nips}, it is therefore also \change{yaw-angle-equivariant} towards features. Then, for the rest part of our attentional feature transformer, we also have:  
\begin{align}
 & \text{LN} ( \text{FFN}(\text{LN}(\text{Conc}(C_s\set{F}, C_s\set{A}))) + \text{LN}(\text{Conc}(C_s\set{F}, C_s\set{A}))) \nonumber \\ 
 &= C_s\text{LN} ( \text{FFN}(\text{LN}(\text{Conc}(\set{F}, \set{A}))) + \text{LN}(\text{Conc}(\set{F}, \set{A}))) \nonumber \\
 &= C_s\set{S} .
 \label{eq:ScaledDotProductAttention}
\end{align}

The transformer module finds the inner connections between different parts of the input feature volume, which exploits the spatial relations of different features in the scene. Similar to humans using distinctive landmarks and their relationship to determine the places, our transformer module also focuses on spatial features and their relationship using the attention mechanism, which boosts the place recognition performance of our method.
\vspace{-0.5em}
\change{\subsection{Global Descriptor Generator}
\label{sec:VLAD}}
NetVLAD was first proposed by Arandjelovic \etalcite{arandjelovic2016netvlad} to tackle image-based place recognition in an end-to-end manner and outperform its non-learning-based counterparts. Uy~\etalcite{uy2018pointnetvlad} transfer it for LiDAR-based place recognition and proof that NetVLAD is permutation invariant, thus suitable for unordered point clouds. \change{However, the original PointNetVLAD is not able to directly generate yaw-angle-invariant global descriptors for the point cloud because the yaw rotation of the point cloud leads to a different input of point coordinates and features. This makes the PointNetVLAD less robust to the rotation or noise of the sensor pose.} 

In this work, we leverage the permutation invariance of NetVLAD together with our \change{yaw-angle-equivariant} features to achieve yaw-rotation invariance and generate \change{yaw-angle-invariant} descriptors. 
We represent the input feature volume with size of $1 \times w \times c$ as a set of 1-D vectors over channel dimension $\set{S} = \set{Z}=\{\v{z}_1, \v{z}_2, ..., \v{z}_i, ..., \v{z}_{w}\}$, where $\v{z}_i$ is the $i$-th vector with size of $c$ channels.
Given that NetVLAD is permutation invariant~\cite{uy2018pointnetvlad}, \change{we have:
\begin{align}
 \text{GDG}(\set{Z} C_s) &= \text{GDG}(\{\v{z}_{w-s}, ..., \v{z}_{w}, \v{z}_1, ..., \v{z}_{w-s-1}\}) \nonumber \\
 &= \text{GDG}(\{\v{z}_1, \v{z}_2, ..., \v{z}_i, ..., \v{z}_{w}\}) \\
 &= \text{GDG}(\set{Z}).
 \label{eq:vlad-permutation-invariant}
\end{align}}
We denote the output descriptor of our \change{global descriptor generator} as $\set{V}$ and \change{we have:
\begin{align}
 \set{V} = \text{GDG}(\set{S}) = \text{GDG}(C_s\set{S}),
 \label{eq:vlad-rotate-inviriant}
\end{align}
which means our proposed global descriptor generator is yaw-angle-invariant.}


\begin{figure}[t]
  \vspace{0.2cm}
  \centering
  \includegraphics[width=1\linewidth]{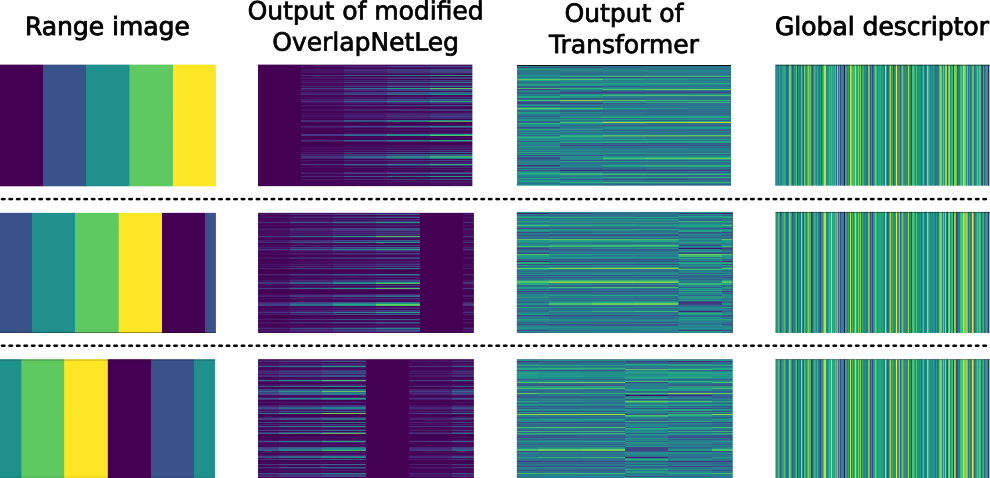}
  \caption{An illustration of our \change{yaw-angle-invariant} descriptors exploiting a toy example. As can be seen that the range image presentation, our range image encoder, and attentional feature transformer are \change{yaw-angle-equivariant}, and our \change{global descriptor generator} generates \change{yaw-angle-invariant} global features.}
  \label{fig:featuremaps}
    \vspace{-0.4cm}
\end{figure}

In~\figref{fig:featuremaps}, we depict a toy example to illustrate that our model can generate \change{yaw-angle-invariant} descriptors with \change{yaw-angle-equivariant} range image representations.
We show at each row a different yaw rotation and the visualizations of the intermediate results of each module of our method. The yaw rotation of point clouds corresponds to the horizontal shift of range image pixels along the width dimension. Supposing the first row is the results of the raw LiDAR data, rotated 0 degree, the second and the third are rotated by 90 and 180 degrees along the yaw angle, which corresponds to the range images shifted by $\frac{1}{4}w$ and $\frac{1}{2}w$ respectively.
As can be seen, the outputs of the \change{yaw-angle-equivariant} OverlapNetLeg and transformer module are also shifted accordingly (shown by the second and the third column). 
Since the \change{global descriptor generator} is \change{yaw-angle-invariant}, it thus generates the same global descriptor in all three cases as shown in the fourth column. 

During online operation, we use the \change{yaw-angle-invariant global} descriptors to represent LiDAR scans and use the Euclidean distance between pairs of descriptors to find the nearest reference places.

\vspace{0.5em}

\subsection{Network Training}
\label{sec:loss}

We follow OverlapNet~\cite{chen2021auro} using overlap to supervise the network rather than directly using ground truth distances. Overlap is a more natural way to describe the similarity between two LiDAR scans than distance. Moreover, the overlap between two LiDAR scans also corresponds to the quality of the following registration, which can be a good criterion for the final metric localization~\cite{chen2021auro}. For a query scan $\set{R}_q$ and a reference scan $\set{R}_r$, we use the ground truth poses to reproject $\set{R}_r$ into the coordinate frame of $\set{R}_q$ and get $\set{R}^{\prime}_r$. The overlap between them is calculated as:
\begin{align}
  O_{\set{R}_q \set{R}_r} & = \frac{\sum_{(u,v)} \mathbb{I}\Big\{\left|\left| \set{R}_q(u,v) - \set{R}^{\prime}_r(u,v) \right|\right| \leq \delta \Big\}} {\min \left( \textrm{valid}(\set{R}_q), \textrm{valid}(\set{R}^{\prime}_r) \right)},  \label{eq:overlap}
\end{align}
where $\mathbb{I}(a)=1$ if $a$ is true and $\mathbb{I}(a)=0$ otherwise, $\text{valid}(\set{R})$ refers to the counts of valid pixels of range image $\set{R}$, and  $\delta$ is the threshold to decide the overlapped pixel.
 
\change{For each training tuple, we utilize one query descriptor $\set{V}_q$, $k_p$ positive descriptors $\{\set{V}_p\}$, and $k_n$ negative descriptors $\{\set{V}_n\}$ to compute lazy triplet loss:
\begin{align}
& \set{L}_T(\set{V}_q,\{\set{V}_p\},\{\set{V}_n\})=    \nonumber \\
& k_p(\alpha+\max_{p}(d(\set{V}_q,\set{V}_p))) 
- \sum_{k_n}(d(\set{V}_q,\set{V}_n)),
\label{eq:tripletloss}
\end{align}}
where $\alpha$ is the margin to avoid negative loss and $d(\cdot)$ is the squared Euclidean distance. We take a pair of scans whose overlap is larger than 0.3 as a positive sample, otherwise a negative sample.
We use the triplet loss to minimize the distance between the query and the hardest positive global descriptors, and maximize the distance between the query and all sampled negative global descriptors. 

\begin{table}[t]
  \centering
  \setlength{\tabcolsep}{3pt}
  \renewcommand\arraystretch{1.1}
  \caption{Statistics of Haomo\;dataset}
  \footnotesize{
\begin{tabular}{lcccccc}
\toprule
Sequence                       & 1-1               & 1-2               & 1-3        & 2-1                & 2-2               \\ \hline
Date                           & 2021.12.8        & 2021.12.8        & 2021.12.8     & 2021.12.28         & 2022.1.13 \\ \hline
N$_\text{scans}$                  & 12500             & 22345             & 13500      & 100887             & 88154             \\ \hline
N$_\text{pos.}$  & 40000  & 40000            & --                & 48000                   & --            \\ \hline
N$_\text{neg.}$ & 60000  & 60000            & --                 & 72000                    & --         \\ \hline
N$_\text{queries}$     & --                & --               & 1350      & --          & 8815              \\ \hline
Length 			   & 2.3\,km & 2.3\,km & 2.3\,km & 11.5\,km & 11.1\,km \\ \hline
Direction                & Same              & Reverse           & --       & Same               & --                \\ \hline
Role                    & Database          & Database          & Query      & Database           & Query             \\ \bottomrule
\multicolumn{6}{p{0.9\linewidth}}{ N$_\text{scans}$, N$_\text{pos.}$, N$_\text{neg.}$ and N$_\text{queries}$ are the numbers of scans, positive samples, negative samples and query scans respectively.} \\
\end{tabular}
  }
  \label{tab:statistics_haomo}
  \vspace{-0.5cm}
\end{table}

\section{Haomo Dataset}
\label{sec:hamodata}

There are several LiDAR-based datasets for autonomous driving \cite{geiger2012cvpr, pandey2011ijrr, wang2019apolloscape}, which can be used for evaluating loop closure detection methods. However, not many of the publicly available datasets show significant repetitive reverse routes for long-term large-scale place recognition of autonomous driving.
In this work, we therefore provide such a new challenging dataset called \emph{Haomo} dataset and released it together with our code to support future research. 
\change{The dataset was collected in urban environments of Beijing by a mobile robot built by HAOMO.AI Technology company equipped with a HESAI PandarXT 32-beam LiDAR sensor, a SENSING-SG2 wide-angle camera, 
and an ASENSING-INS570D RTK GNSS.} 
There are currently five sequences in Haomo dataset as listed in~\tabref{tab:statistics_haomo}. Sequences 1-1 and 1-2 are collected from the same route in 8th December 2021 with opposite driving direction. An additional sequence 1-3 from the same route is utilized as the online query with respect to both 1-1 and 1-2 respectively to evaluate place recognition performance of forward and reverse driving.  Sequences 2-1 and 2-2 are collected along a much longer route from the same direction, but on different dates, 2-1 on 28th December 2021 and 2-2 on 13th January 2022, where the old one is used as a database while the newer one is used as query. The two sequences are for evaluating the performance for large-scale long-term place recognition. 
 



\section{Experimental Evaluation}
\label{sec:exp}
The experimental evaluation is designed to showcase the performance of our approach and to evaluate the claims that our approach is able to:
(i)~detect loop closure candidates for SLAM using only LiDAR data without any other  information, and generalize well into the different environments without fine-tuning,
(ii)~achieve long-term place recognition on our Haomo dataset in outdoor large-scale environments,
(iii)~recognize places with changing viewpoints exploiting the proposed yaw-angle-invariant descriptors,
(iv)~run faster than most state-of-the-art place recognition methods.

\subsection{Implementation and Experimental Setup}
We use three different datasets to evaluate our method, including KITTI dataset~\cite{geiger2012cvpr} collected in Germany with a 64-beam LiDAR sensor, Ford Campus dataset~\cite{pandey2011ijrr} collected in U.S. with a 64-beam LiDAR sensor, and our Haomo dataset collected in China with a 32-beam LiDAR sensor. Following OverlapNet~\cite{chen2021auro}, we use range images of size $1\times64\times 900$ for 64-beam LiDAR data of KITTI and Ford Campus datasets, and range images with size of $1\times32\times900$ for our Haomo dataset with 32-beam LiDAR data. 
For our transformer module, we set the embedding dimension $d_{model}=256$, the number of heads $n_{head}=4$, and the intermediate dimension of the feed-forward layer $d_\mathit{ffn}=1024$ . \change{We do not utilize dropout to achieve yaw-angle-equivariant}.
For NetVLAD, we set the intermediate feature dimension $d_{inter}=1024$, the output feature dimension $d_{output}=256$, and the number of clusters $d_{K}=64$ . The output of our \change{global descriptor generator} is a vector of size $256$.
For calculating overlap ground truth by \eqref{eq:overlap}, we set $\delta=1$ for KITTI odometry benchmark with provided 64-beam LiDAR scans, $\delta=1.2$ for our Haomo dataset depending on the density of the points.
We set $k_p=6$, $k_n=6$, and $\alpha=0.5$ for the triplet loss.


\subsection{Evaluation for Loop Closure Detection}
\label{lcd}

\begin{table}[t]
  \centering
  \setlength{\tabcolsep}{1.6pt}
  \renewcommand\arraystretch{1.1}
  \caption{Comparison of loop closure detection performance}
  \footnotesize{
\begin{tabular}{l|l|C{1cm}C{1cm}C{1cm}C{1cm}}
\toprule
Dataset                      & Approach                   & AUC             & F1max           & Recall @1        & Recall @1\%      \\ \hline
\multirow{5}{*}{KITTI}       & Histogram~\cite{roehling2015iros}               & 0.826          & 0.825          & 0.738          & 0.871          \\ \cline{2-6} 
                             & Scan Context~\cite{kim2018scan}               & 0.836          & 0.835          & 0.820          & 0.869          \\ \cline{2-6} 
                             & LiDAR Iris~\cite{wang2020lidar}                 & 0.843          & 0.848          & 0.835          & 0.877          \\ \cline{2-6} 
                             & PointNetVLAD~\cite{uy2018pointnetvlad}               & 0.856          & 0.846          & 0.776          & 0.845          \\ \cline{2-6} 
                             & OverlapNet~\cite{chen2021auro}                 & 0.867            & 0.865            & 0.816             & 0.908             \\ \cline{2-6} 
                             & \change{NDT-Transformer-P~\cite{zhou2021ndt}}                 & \change{0.855}            & \change{0.853}            & \change{0.802}             & \change{0.869}             \\ \cline{2-6}                              

                             & \change{MinkLoc3D~\cite{Komorowski2021wacv}}                 & \change{0.894}            & \change{0.869}            & \change{0.876}             & \change{0.920}             \\ \cline{2-6} 

                             & \textbf{Ours}                       & \textbf{0.907} & \textbf{0.877} & \textbf{0.906} & \textbf{0.964} \\ \midrule                            
\multirow{5}{1cm}{Ford Campus}       & Histogram~\cite{roehling2015iros}               & 0.841          & 0.800          & 0.812          & 0.897          \\ \cline{2-6} 
                             & Scan Context~\cite{kim2018scan}                 & 0.903          & 0.842          & 0.878          & \textbf{0.958} \\ \cline{2-6} 
                            & LiDAR Iris~\cite{wang2020lidar}                    & 0.907          & 0.842          & 0.849          & 0.937         \\ \cline{2-6} 
                              & PointNetVLAD~\cite{uy2018pointnetvlad}               & 0.872          & 0.830          & 0.862         & 0.938          \\ \cline{2-6} 
                             & OverlapNet~\cite{chen2021auro}                 & 0.854            & 0.843            & 0.857             & 0.932             \\ \cline{2-6} 
                             & \change{NDT-Transformer-P~\cite{zhou2021ndt}}                 & \change{0.835}            & \change{0.850}            & \change{0.900}             & \change{0.927}             \\ \cline{2-6}                              
                             & \change{MinkLoc3D~\cite{Komorowski2021wacv}}                 & \change{0.871}            & \change{0.851}            & \change{0.878}             & \change{0.942}             \\ \cline{2-6} 
                             & \textbf{Ours}                       & \textbf{0.923} & \textbf{0.856} & \textbf{0.914} & 0.954          \\ \bottomrule
\end{tabular}
  }
  \label{tab:loopclosuremetric}
  \vspace{-0.2cm}
\end{table}

The first experiment supports our claim that our approach detects loop closure candidates for SLAM using only LiDAR data without any other information, and generalizes well into the different environments without fine-tuning.
Following the experimental setup of OverlapNet \cite{chen2021auro}, we train and evaluate our approach and other learning-based baseline methods on the KITTI Odometry Benchmark~\cite{geiger2012cvpr}.
We use sequences~$03$--$10$ for training, sequence~$02$ for validation, and sequence~$00$ for evaluation. \change{We regard two scans as a loop closure if their overlap value is larger than $0.3$.}
To evaluate the generalization ability of our method, we also test it on sequence~$00$ of the Ford Campus dataset~\cite{pandey2011ijrr}. Note that we did not \change{train} our approach on the Ford Campus dataset, which shows the generalization ability of our method.

\change{Since the semantics are not always available for the different datasets, we compare our method with existing methods which do not utilize semantic information, including Histogram \cite{roehling2015iros}, Scan Context \cite{kim2018scan} with augmentation, LiDAR Iris \cite{wang2020lidar}, PointNetVLAD \cite{uy2018pointnetvlad}, OverlapNet~(Delta-Geo-Only)~\cite{chen2021auro}, NDT-Transformer-P~(4096 cells) \cite{zhou2021ndt}, and MinkLoc3D \cite{Komorowski2021wacv}}.
Loop closure detection during SLAM takes previous scans as the database, excluding the nearby 100 scans to avoid detecting the most recent scans.
We utilize AUC, F1 max scores, recall@1 and recall@1\% for evaluation and the results are shown in \tabref{tab:loopclosuremetric}.
As can be seen, our method outperforms all the baseline methods on the KITTI dataset.
Using our network pre-trained on KITTI directly on an unseen environment, Ford Campus dataset, without any fine-tuning, our proposed method still outperforms other methods, and only for recall@1\%, our method is on par with the state-of-the-art non-learning-based method Scan Context~\cite{kim2018scan}, which shows the good generalization ability of our method.

\subsection{Evaluation for Place Recognition}

In the second experiment, we investigate the long-term place recognition of our method on our Haomo dataset in outdoor large-scale environments with a different type of LiDAR sensor.
The difference between loop closure detection and place recognition is that for place recognition the database/map is usually given and we need to find the location of the vehicle within the database/map.
For loop closing in contrast, the database grows during operation.
Therefore, we train our approach and other learning-based baseline methods on the database sequences and evaluate all the methods using the same new query sequences that were not parts of the database.
We have three challenges in Haomo dataset. From easy to difficult, the first is a short-term same-direction challenge, sequence 1-1 as database and 1-3 for query. The second is the short-term inverse-direction challenge, sequence 1-2 as database and 1-3 for query. The last is a long-term large-scale challenge, sequence 2-1 as database and 2-2 for query.
We use database sequences for training too. As shown in~\tabref{tab:statistics_haomo}, we firstly downsample the database and for each sampled laser scan, we calculate the overlaps between this scan to all other scans. 
\change{For testing, we take one scan sampled from every ten scans in query sequences, and acquire its ground truth reference scans with overlap values larger than $0.3$.}

\begin{figure}[t]
  \centering
  \includegraphics[width=0.9\linewidth]{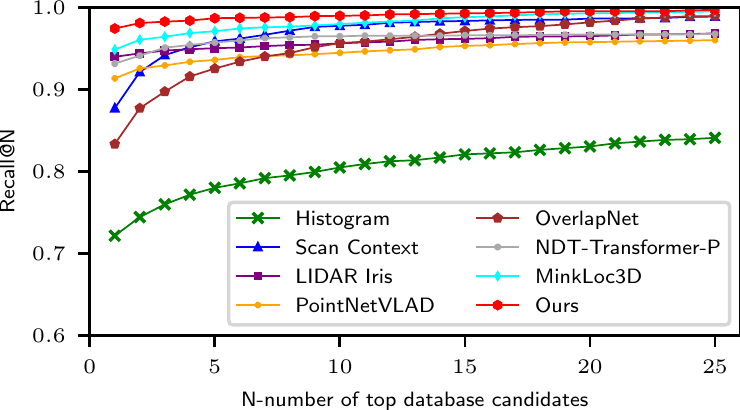}
  \caption{\change{Place recognition results using Haomo dataset sequence 1-3 as query and 1-1 as database (driving forward).}}
  \label{fig:haomo_seq00_new}
    \vspace{-0.2cm}
\end{figure}

\begin{figure}[t]
  \centering
  \includegraphics[width=0.9\linewidth]{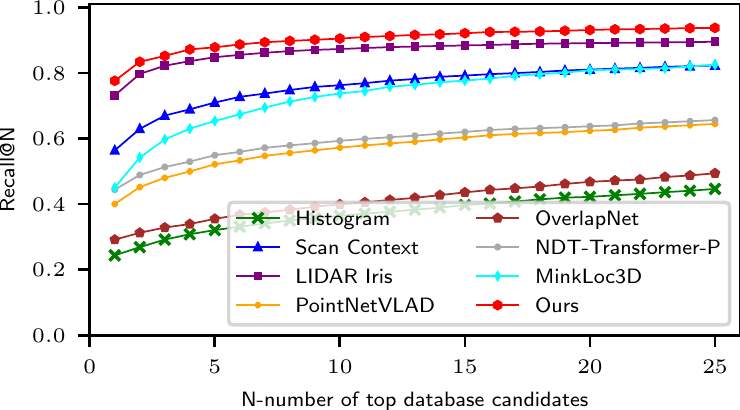}
  \caption{\change{Place recognition results using Haomo dataset sequence 1-3 as query and 1-2 as database (driving in reverse).}}
  \label{fig:haomo_seq00_reverse_new}
    \vspace{-0.2cm}
\end{figure}

\begin{figure}[t]
  \centering
  \includegraphics[width=0.9\linewidth]{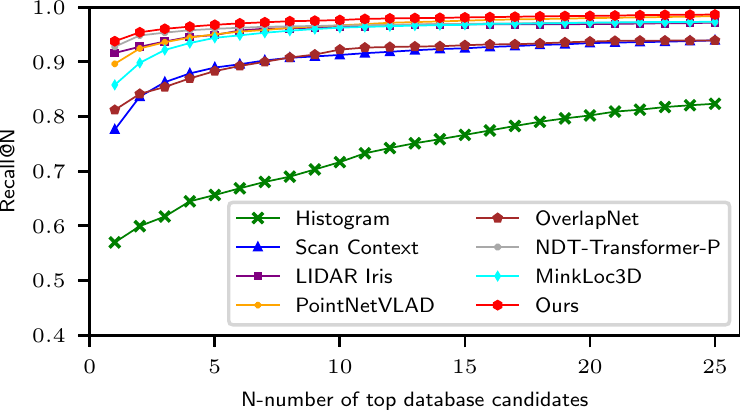}
  \caption{\change{Place recognition results using Haomo dataset sequence 2-2 as query and 2-1 as database (driving in large-scale environments).}}
  \label{fig:haomo_seq01_new}
    \vspace{-0.2cm}
\end{figure}

We utilize recall@N to evaluate the performance of algorithms on Haomo dataset as the most large-scale place recognition approaches do. \change{We consider one query as a successful loop closure once we find one of the true references.}
The results of the first two challenges are shown in \figref{fig:haomo_seq00_new} and \figref{fig:haomo_seq00_reverse_new}. As can be seen, our method outperforms other methods on these two challenges, especially for the reverse driving one due to the \change{yaw-angle-invariant} architecture design of our method. 
Although Scan Context and Histogram are also robust to pure rotation, our method outperforms them, since in real application there is not only pure yaw rotation but also changes in locations and environment appearance due to the dynamic objects. Our method is more robust to all challenging conditions compared to baseline methods.
The evaluation on long-term large-scale sequences 2-2 and 2-1 are illustrated in \figref{fig:haomo_seq01_new}, and our method also keeps its superiority on large-scale long-term datasets, which shows the ability of our method to be used for real autonomous driving applications.

\begin{figure}[t]
  \centering
  \includegraphics[width=0.9\linewidth]{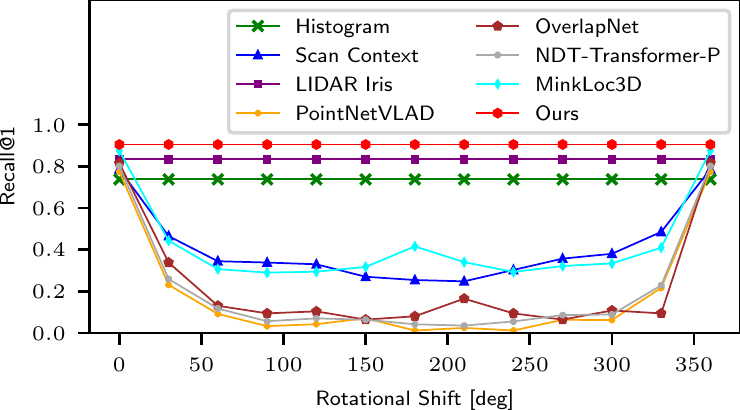}
  \caption{\change{yaw-angle-invariant test.}}
  \label{fig:rit}
    \vspace{-0.2cm}
\end{figure}

\change{\subsection{Study on Yaw-Angle-Invariance}}
The third experiment investigates the \change{yaw-angle-invariance} of our approach. The results support the claim that our method generates \change{yaw-angle-invariant} descriptors and can recognize places with changing viewpoints.
For that, we rotate each query scan of the KITTI dataset along the yaw-axis in steps of 30 degrees and search the places with respect to the same original database. In this experiment, we test all baselines and use Recall@1 as the evaluation metric. As illustrated in~\figref{fig:rit}, our proposed OverlapTransformer is not influenced by pure rotation along the yaw-axis, and maintains the best performance compared to the baseline methods. \change{On the contrary, PointNetVLAD, Scan Context, OverlapNet, NDT-Transformer-P and MinkLoc3D are all affected by rotation. PointNetVLAD, OverlapNet, and NDT-Transformer-P lose efficacy quickly for increasing yaw angle discrepancies. Here, we only use OverlapNet with the head of the similarity estimation for a fair comparison. Scan Context loses efficacy to some extent but is better than PointNetVLAD since the ring key representing occupancy ratio is yaw-angle-invariant. MinkLoc3D has the similar performance as Scan Context.} 
The LiDAR Iris and histogram-based method are also \change{yaw-angle-invariant} as our method but have lower recall@1 than ours. LiDAR Iris achieves \change{yaw-angle-invariant} by rotating its features multiple times and choosing the best scored one, which makes it very slow as shown in~\secref{sec:runtime}. 
Our OverlapTransformer outperforms other baseline methods significantly due to our devised \change{yaw-angle-invariant} model exploiting \change{yaw-angle-equivariant} range images.

\change{\subsection{Ablation Study on Transformer Module}} 
\label{sec:ablation}



\change{This ablation study validates the effectiveness of the transformer module (TM) we used between our range image encoder (RIE) and global descriptor generator (GDG). We use Haomo dataset sequence 1-2 as database and 1-3 for query to compare 5 different setups, including RIE+GDG, RIE+Conv+GDG, RIE+1TM+GDG, RIE+3TM+GDG, and RIE+6TM+GDG. RIE+GDG uses no transformer module. RIE+Conv+GDG replaces the transformer module with two 1$\times$1 convolution layers without changing the channel numbers. RIE+1/3/6TM+GDG uses 1, 3, 6 TMs respectively. As shown in~\tabref{tab:ablation}, RIE+3TM+GDG and RIE+1TM+GDG outperform the other setups for place recognition. The results show that one transformer block already increases the performance significantly, while more transformer blocks lead to lower efficiency. When using more than 3 transformer blocks, the performance even decreases slightly. More transformer blocks might need more training data and time to obtain good performance. Thus, we only use one transformer block.}

\change{
\begin{table}[t]
  \centering
  \setlength{\tabcolsep}{3pt}
  \renewcommand\arraystretch{1.1}
  \caption{\change{Ablation study on the transformer module}}
  \footnotesize{
\begin{tabular}{L{60pt}|C{45pt}|C{35pt}C{35pt}C{36pt}}
\toprule
Network    & Runtime [ms]     & Recall@1               & Recall@5            & Recall@20                 \\ \hline
RIE+GDG    & 0.86         & 0.642               & 0.750            & 0.841                  \\ \hline
RIE+Conv+GDG   & 0.92         & 0.698         & 0.841        & 0.923         \\ \hline
RIE+1TM+GDG      & 1.37              & 0.776             & \textbf{0.878}             & 0.931                      \\ \hline
RIE+3TM+GDG   & 2.39      & \textbf{0.788}  & 0.875            & \textbf{0.940}                  \\ \hline
RIE+6TM+GDG    & 3.95     & 0.741  & 0.852            & 0.913                
 \\ \bottomrule
\end{tabular}
  }
  \label{tab:ablation}
  \vspace{-0.4cm}
\end{table}
}

\subsection{Runtime}
\label{sec:runtime}

\begin{table}[t]
  \centering
  \vspace{0.2cm}
  \setlength{\tabcolsep}{2pt}
  \renewcommand\arraystretch{1.1}
  \caption{Comparison of runtime with \change{state-of-the-art methods}}
  \footnotesize{
\begin{tabular}{l|l|C{2cm}C{2cm}}
\toprule
\multicolumn{2}{c|}{Approach}                    & Descriptor Extraction {[}ms{]} & Searching {[}ms{]}     \\ \hline
\multirow{3}{1cm}{Hand crafted}   & Histogram~\cite{roehling2015iros}     & \textbf{1.07}                  & \textbf{0.46}               \\ \cline{2-4} 
                                & Scan Context~\cite{kim2018scan}  & 57.95                          & 492.63                    \\ \cline{2-4} 
                                & LiDAR Iris~\cite{wang2020lidar}    & 7.13                           & 9315.16                    \\ \hline
\multirow{5}{1cm}{Learning based} & PointNetVLAD~\cite{uy2018pointnetvlad}  & 13.87                          & 1.43                     \\ \cline{2-4} 
                                & OverlapNet~\cite{chen2021auro}    & 4.85                           & 3233.30                  \\ \cline{2-4} 
                                & \change{NDT-Transformer-P~\cite{zhou2021ndt}}    & \change{15.73}                          & \change{0.49}                  \\ \cline{2-4} 

                                & \change{MinkLoc3D~\cite{Komorowski2021wacv}}    & \change{15.94}                          & \change{8.10}                \\ \cline{2-4} 
                                & Ours                     & \textbf{1.37}                  & \textbf{0.44}         \\ \bottomrule
\end{tabular}
  }
  \label{tab:runtime}
  \vspace{-0.5cm}
\end{table}

The experiment evaluates the runtime requirements of our method. \change{It supports our last claim that our method runs faster than most of the state-of-the-art place recognition methods and that it can run at 730\,Hz, \ie, much faster than the scanning rate.}
We compare the runtime of our OverlapTransformer with all baseline methods. We conduct all experiments on a system with an Intel i7-11700K CPU and an Nvidia RTX 3070 GPU. We run all the methods to find top-1 candidates for one query scan with respect to a database consisting of 2000 reference scans. We report the averaged results over ten experiments.
As shown in~\tabref{tab:runtime}, we count the runtime separately for descriptor generation and searching.
For the time cost by generating the descriptor, we take also the LiDAR data preprocessing into consideration. 
\change{As can be seen, for descriptor generation, our method is the fastest method among the state-of-the-art learning-based methods with 1.37\,ms for each scan, and slightly slower than the pure geometric histogram-based method.
PointNetVLAD, NDT-Transformer-P, and MinkLoc3D need an extra downsampling process for raw point cloud, which is accelerated by GPU. OverlapNet needs to estimate the normals. 
For the time cost by searching, our method outperforms all baselines including the non-learning-based methods. Since Histogram, PointNetVLAD, NDT-Transformer-P, MinkLoc3D and our method do not need an ad-hoc function to compute the similarity of descriptors, FAISS library~\cite{johnson2019billion} is used to accelerate the searching.}
\change{Note that, PointNetVLAD, NDT-Transformer-P and MinkLoc3D generate descriptors with the same size ($1 \times 256$) as ours, but consume more time. The reason could be that the descriptors generated by our method are more descriptive than those by PointNetVLAD, NDT-Transformer-P, and MinkLoc3D, and thus need less time for searching.}

\section{Conclusion}
\label{sec:conclusion}

\change{In this paper, we presented a novel approach for LiDAR-based place recognition with a runtime less than 2\,ms and outperforms the state-of-the-art methods in terms of place recognition performance. Our approach utilizes a lightweight network with a Transformer attention mechanism to generate yaw-angle-invariant descriptors, which allows us to handle challenging place recognition effectively and efficiently.} We trained our method on the KITTI dataset and evaluated it on both, the KITTI and the Ford Campus dataset for loop closure detection, which shows a solid generalization capability. For further evaluating the place recognition ability on long time spans, reverses driving, and large-scale scenes, we developed a new challenging Haomo dataset and conducted extensive evaluations with multiple baseline approaches on it. The experimental results suggest that our method outperforms the other state-of-the-art methods in different challenging environments in terms of recognition performance and speed.



\bibliographystyle{IEEEtran}

\footnotesize{
\bibliography{glorified,new}}

\begin{thebibliography}{10}
\providecommand{\url}[1]{#1}
\csname url@rmstyle\endcsname
\providecommand{\newblock}{\relax}
\providecommand{\bibinfo}[2]{#2}
\providecommand\BIBentrySTDinterwordspacing{\spaceskip=0pt\relax}
\providecommand\BIBentryALTinterwordstretchfactor{4}
\providecommand\BIBentryALTinterwordspacing{\spaceskip=\fontdimen2\font plus
\BIBentryALTinterwordstretchfactor\fontdimen3\font minus
  \fontdimen4\font\relax}
\providecommand\BIBforeignlanguage[2]{{%
\expandafter\ifx\csname l@#1\endcsname\relax
\typeout{** WARNING: IEEEtran.bst: No hyphenation pattern has been}%
\typeout{** loaded for the language `#1'. Using the pattern for}%
\typeout{** the default language instead.}%
\else
\language=\csname l@#1\endcsname
\fi
#2}}

\bibitem{arandjelovic2016netvlad}
R.~Arandjelovic, P.~Gronat, A.~Torii, T.~Pajdla, and J.~Sivic, ``Netvlad: Cnn
  architecture for weakly supervised place recognition,'' in \emph{Proc.~of the
  IEEE/CVF Conf.~on Computer Vision and Pattern Recognition (CVPR)}, 2016.

\bibitem{hausler2021patch}
S.~Hausler, S.~Garg, M.~Xu, M.~Milford, and T.~Fischer, ``Patch-netvlad:
  Multi-scale fusion of locally-global descriptors for place recognition,'' in
  \emph{Proc.~of the IEEE/CVF Conf.~on Computer Vision and Pattern Recognition
  (CVPR)}, 2021.

\bibitem{lowry2016tro}
S.~Lowry, N.~Sunderhauf, P.~Newman, J.~Leonard, D.~Cox, P.~Corke, and
  M.~Milford, ``Visual place recognition: A survey,'' \emph{IEEE Trans.~on
  Robotics (TRO)}, vol.~32, no.~1, pp. 1--19, 2016.

\bibitem{vysotska2016ral}
O.~Vysotska and C.~Stachniss, ``{Lazy Data Association For Image Sequences
  Matching Under Substantial Appearance Changes},'' \emph{IEEE Robotics and
  Automation Letters (RA-L)}, vol.~1, no.~1, pp. 213--220, 2016.

\bibitem{vysotska2019ral}
------, ``{Effective Visual Place Recognition Using Multi-Sequence Maps},''
  \emph{IEEE Robotics and Automation Letters (RA-L)}, vol.~4, pp. 1730--1736,
  2019.

\bibitem{chen2020rss}
X.~Chen, T.~L\"abe, A.~Milioto, T.~R\"ohling, O.~Vysotska, A.~Haag, J.~Behley,
  and C.~Stachniss, ``{OverlapNet: Loop Closing for LiDAR-based SLAM},'' in
  \emph{Proc.~of Robotics: Science and Systems (RSS)}, 2020.

\bibitem{kim2018scan}
G.~Kim and A.~Kim, ``Scan context: Egocentric spatial descriptor for place
  recognition within 3d point cloud map,'' in \emph{Proc.~of the IEEE/RSJ
  Intl.~Conf.~on Intelligent Robots and Systems (IROS)}, 2018.

\bibitem{uy2018pointnetvlad}
M.~A. Uy and G.~H. Lee, ``Pointnetvlad: Deep point cloud based retrieval for
  large-scale place recognition,'' in \emph{Proc.~of the IEEE/CVF Conf.~on
  Computer Vision and Pattern Recognition (CVPR)}, 2018.

\bibitem{wang2020lidar}
Y.~Wang, Z.~Sun, C.-Z. Xu, S.~E. Sarma, J.~Yang, and H.~Kong, ``Lidar iris for
  loop-closure detection,'' in \emph{Proc.~of the IEEE/RSJ Intl.~Conf.~on
  Intelligent Robots and Systems (IROS)}, 2020.

\bibitem{chen2019iros}
X.~Chen, A.~Milioto, E.~Palazzolo, P.~Giguère, J.~Behley, and C.~Stachniss,
  ``{SuMa++: Efficient LiDAR-based Semantic SLAM},'' in \emph{Proc.~of the
  IEEE/RSJ Intl.~Conf.~on Intelligent Robots and Systems (IROS)}, 2019.

\bibitem{chen2021auro}
X.~Chen, T.~L\"abe, A.~Milioto, T.~R\"ohling, J.~Behley, and C.~Stachniss,
  ``{OverlapNet: A Siamese Network for Computing LiDAR Scan Similarity with
  Applications to Loop Closing and Localization},'' \emph{Autonomous Robots},
  vol.~46, pp. 61--81, 2021.

\bibitem{chen2020iros}
X.~Chen, T.~L\"abe, L.~Nardi, J.~Behley, and C.~Stachniss, ``{Learning an
  Overlap-based Observation Model for 3D LiDAR Localization},'' in
  \emph{Proc.~of the IEEE/RSJ Intl.~Conf.~on Intelligent Robots and Systems
  (IROS)}, 2020.

\bibitem{chen2021icra}
X.~Chen, I.~Vizzo, T.~L\"abe, J.~Behley, and C.~Stachniss, ``{Range Image-based
  LiDAR Localization for Autonomous Vehicles},'' in \emph{Proc.~of the IEEE
  Intl.~Conf.~on Robotics \& Automation (ICRA)}, 2021.

\bibitem{he2016iros}
L.~He, X.~Wang, and H.~Zhang, ``{M2DP: A Novel 3D Point Cloud Descriptor and
  Its Application in Loop Closure Detection},'' in \emph{Proc.~of the IEEE/RSJ
  Intl.~Conf.~on Intelligent Robots and Systems (IROS)}, 2016.

\bibitem{roehling2015iros}
T.~R{\"o}hling, J.~Mack, and D.~Schulz, ``{A Fast Histogram-Based Similarity
  Measure for Detecting Loop Closures in 3-D LIDAR Data},'' in \emph{Proc.~of
  the IEEE/RSJ Intl.~Conf.~on Intelligent Robots and Systems (IROS)}, 2015, pp.
  736--741.

\bibitem{steder2010irosws}
B.~Steder, R.~Rusu, K.~Konolige, and W.~Burgard, ``{NARF}: {3D} range image
  features for object recognition,'' in \emph{Proc.~of~the~IROS Workshop on
  Defining and Solving Realistic Perception Problems in Personal Robotics},
  2010.

\bibitem{vaswani2017nips}
A.~Vaswani, N.~Shazeer, N.~Parmar, J.~Uszkoreit, L.~Jones, A.~N. Gomez,
  {\L}.~Kaiser, and I.~Polosukhin, ``Attention is all you need,'' in
  \emph{Proc.~of the Advances in Neural Information Processing Systems (NIPS)},
  2017.

\bibitem{jegou2010cvpr}
H.~J{\'e}gou, M.~Douze, C.~Schmid, and P.~P{\'e}rez, ``{Aggregating local
  descriptors into a compact image representation},'' in \emph{Proc.~of the
  IEEE/CVF Conf.~on Computer Vision and Pattern Recognition (CVPR)}, 2010.

\bibitem{cop2018delight}
K.~P. Cop, P.~V. Borges, and R.~Dub{\'e}, ``Delight: An efficient descriptor
  for global localisation using lidar intensities,'' in \emph{Proc.~of the IEEE
  Intl.~Conf.~on Robotics \& Automation (ICRA)}, 2018.

\bibitem{wang2020intensity}
H.~Wang, C.~Wang, and L.~Xie, ``Intensity scan context: Coding intensity and
  geometry relations for loop closure detection,'' in \emph{Proc.~of the IEEE
  Intl.~Conf.~on Robotics \& Automation (ICRA)}, 2020.

\bibitem{dube2017segmatch}
R.~Dub{\'e}, D.~Dugas, E.~Stumm, J.~Nieto, R.~Siegwart, and C.~Cadena,
  ``Segmatch: Segment based place recognition in 3d point clouds,'' in
  \emph{Proc.~of the IEEE Intl.~Conf.~on Robotics \& Automation (ICRA)}, 2017.

\bibitem{vidanapathirana2021locus}
K.~Vidanapathirana, P.~Moghadam, B.~Harwood, M.~Zhao, S.~Sridharan, and
  C.~Fookes, ``Locus: Lidar-based place recognition using spatiotemporal
  higher-order pooling,'' in \emph{Proc.~of the IEEE Intl.~Conf.~on Robotics \&
  Automation (ICRA)}, 2021.

\bibitem{liu2019lpd}
Z.~Liu, S.~Zhou, C.~Suo, P.~Yin, W.~Chen, H.~Wang, H.~Li, and Y.-H. Liu,
  ``Lpd-net: 3d point cloud learning for large-scale place recognition and
  environment analysis,'' in \emph{Proc.~of the IEEE/CVF Intl.~Conf.~on
  Computer Vision (ICCV)}, 2019.

\bibitem{qi2017pointnet}
C.~R. Qi, H.~Su, K.~Mo, and L.~J. Guibas, ``Pointnet: Deep learning on point
  sets for 3d classification and segmentation,'' in \emph{Proc.~of the IEEE/CVF
  Conf.~on Computer Vision and Pattern Recognition (CVPR)}, 2017.

\bibitem{Komorowski2021wacv}
J.~Komorowski, ``{MinkLoc3D: Point Cloud Based Large-Scale Place
  Recognition},'' in \emph{Proc.~of the IEEE Winter Conf.~on Applications of
  Computer Vision (WACV)}, 2021.

\bibitem{vid2022icra}
K.~Vidanapathirana, M.~Ramezani, P.~Moghadam, S.~Sridharan, and C.~Fookes,
  ``{LoGG3D-Net: Locally Guided Global Descriptor Learning for 3D Place
  Recognition},'' in \emph{Proc.~of the IEEE Intl.~Conf.~on Robotics \&
  Automation (ICRA)}, 2022.

\bibitem{xia2021soe}
Y.~Xia, Y.~Xu, S.~Li, R.~Wang, J.~Du, D.~Cremers, and U.~Stilla, ``Soe-net: A
  self-attention and orientation encoding network for point cloud based place
  recognition,'' in \emph{Proc.~of the IEEE/CVF Conf.~on Computer Vision and
  Pattern Recognition (CVPR)}, 2021.

\bibitem{zhou2021ndt}
Z.~Zhou, C.~Zhao, D.~Adolfsson, S.~Su, Y.~Gao, T.~Duckett, and L.~Sun,
  ``Ndt-transformer: Large-scale 3d point cloud localisation using the normal
  distribution transform representation,'' in \emph{Proc.~of the IEEE
  Intl.~Conf.~on Robotics \& Automation (ICRA)}, 2021.

\bibitem{hui2021iccv}
L.~Hui, H.~Yang, M.~Cheng, J.~Xie, and J.~Yang, ``{Pyramid Point Cloud
  Transformer for Large-Scale Place Recognition},'' in \emph{Proc.~of the
  IEEE/CVF Intl.~Conf.~on Computer Vision (ICCV)}, 2021.

\bibitem{kong2020semantic}
X.~Kong, X.~Yang, G.~Zhai, X.~Zhao, X.~Zeng, M.~Wang, Y.~Liu, W.~Li, and
  F.~Wen, ``Semantic graph based place recognition for 3d point clouds,'' in
  \emph{Proc.~of the IEEE/RSJ Intl.~Conf.~on Intelligent Robots and Systems
  (IROS)}, 2020.

\bibitem{li2021ssc}
L.~Li, X.~Kong, X.~Zhao, T.~Huang, W.~Li, F.~Wen, H.~Zhang, and Y.~Liu, ``{Ssc:
  Semantic scan context for large-scale place recognition},'' in \emph{Proc.~of
  the IEEE/RSJ Intl.~Conf.~on Intelligent Robots and Systems (IROS)}, 2021.

\bibitem{cramariuc2021semsegmap}
A.~Cramariuc, F.~Tschopp, N.~Alatur, S.~Benz, T.~Falck, M.~Br{\"u}hlmeier,
  B.~Hahn, J.~Nieto, and R.~Siegwart, ``Semsegmap--3d segment-based semantic
  localization,'' in \emph{Proc.~of the IEEE/RSJ Intl.~Conf.~on Intelligent
  Robots and Systems (IROS)}, 2021.

\bibitem{geiger2012cvpr}
A.~Geiger, P.~Lenz, and R.~Urtasun, ``{Are we ready for Autonomous Driving? The
  KITTI Vision Benchmark Suite},'' in \emph{Proc.~of the IEEE Conf.~on Computer
  Vision and Pattern Recognition (CVPR)}, 2012.

\bibitem{pandey2011ijrr}
G.~Pandey, J.~McBride, and R.~Eustice, ``{Ford campus vision and lidar data
  set},'' \emph{Intl.~Journal~of Robotics Research (IJRR)}, vol.~30, no.~13,
  pp. 1543--1552, 2011.

\bibitem{wang2019apolloscape}
P.~Wang, X.~Huang, X.~Cheng, D.~Zhou, Q.~Geng, and R.~Yang, ``The apolloscape
  open dataset for autonomous driving and its application,'' \emph{IEEE
  Trans.~on Pattern Analalysis and Machine Intelligence (TPAMI)}, 2019.

\bibitem{johnson2019billion}
J.~Johnson, M.~Douze, and H.~J{\'e}gou, ``Billion-scale similarity search with
  gpus,'' \emph{IEEE Trans.~on Big Data}, vol.~7, no.~3, pp. 535--547, 2019.

\end{thebibliography}

\end{document}